\documentclass{article} 
\usepackage{iclr2022_conference,times}

\usepackage{amsmath,amsfonts,bm}

\def\eqref#1{equation~\ref{#1}}

\def\1{\bm{1}}

\DeclareMathAlphabet{\mathsfit}{\encodingdefault}{\sfdefault}{m}{sl}
\SetMathAlphabet{\mathsfit}{bold}{\encodingdefault}{\sfdefault}{bx}{n}

\DeclareMathOperator*{\argmax}{arg\,max}

\usepackage{hyperref}
\usepackage{url}
\usepackage{multirow}
\usepackage{graphicx}
\usepackage{amsmath}
\usepackage{booktabs}
\usepackage{siunitx}
\usepackage{subcaption}
\usepackage{tikz} 
\usepackage{fancyvrb}
\usepackage{cprotect}
\usetikzlibrary{arrows.meta}
\usetikzlibrary{arrows.meta,calc,decorations.markings,math,arrows.meta}
\usepackage{bbm}
\DeclareMathOperator*{\argtopk}{arg\,topK}
\usepackage[ruled,linesnumbered,noend]{algorithm2e} 

\usepackage[T1]{fontenc}

\definecolor{diffstart}{named}{gray}
\definecolor{diffincl}{named}{teal}
\definecolor{diffrem}{named}{red}
\usepackage{listings}
\lstdefinestyle{diff}{
	morecomment=[f][\color{diffstart}]{?},
	morecomment=[f][\color{diffincl}]{+\ },
	morecomment=[f][\color{diffrem}]{-\ },
}
\lstdefinestyle{diffjava} {basicstyle =\footnotesize\ttfamily, language=java,style=diff,frame=lines,  
	breaklines=true,
	postbreak=\mbox{\textcolor{red}{$\hookrightarrow$}\space}
}
\title{Fix Bugs with Transformer through a Neural-Symbolic Edit Grammar}

\author{Yaojie Hu\thanks{Work completed while at AWS AI}\\
Department of Computer Science\\
Iowa State University\\
\texttt{jhu@iastate.edu} \\
\And
Xingjian Shi, Qiang Zhou, Lee Pike\\
AWS AI\\
{\texttt{\{xjshi, zhouqia, leepike\}@amazon.com}}
}

\iclrfinalcopy 

\begin{document}
\maketitle
\begin{abstract}
	We introduce NSEdit (neural-symbolic edit), a novel Transformer-based code repair method. Given only the source code that contains bugs, NSEdit predicts an editing sequence that can fix the bugs. The edit grammar is formulated as a regular language, and the Transformer uses it as a neural-symbolic scripting interface to generate editing programs. We modify the Transformer and add a pointer network to select the edit locations. An ensemble of rerankers are trained to re-rank the editing sequences generated by beam search. We fine-tune the rerankers on the validation set to reduce over-fitting. NSEdit is evaluated on various code repair datasets and achieved a new state-of-the-art accuracy ($24.04\%$) on the Tufano small dataset of the CodeXGLUE benchmark. NSEdit performs robustly when programs vary from packages to packages and when buggy programs are concrete. We conduct detailed analysis on our methods and demonstrate the effectiveness of each component.
\end{abstract}

\section{Introduction}

Neural networks pretrained on source code \citep{feng2020codebert, kanade2020learning,guo2020graphcodebert,ahmad2021unified} are bringing substantial gains on many code understanding tasks such as code classification, code search, and code completion \citep{lu2021codexglue}.  However, code repair remains challenging because it requires the model to have robust syntactic and semantic understanding of a given program even when it contains bugs. The difference between a buggy program and its fixed version often lies in small details that must be fixed exactly, which further requires the model to have precise knowledge about programming.

Code repair with large language models often formulate the problem as a Neural Machine Translation problem (NMT), where the input buggy code is ``translated'' into its fixed version as the output. We categorize existing work based on two design factors: the \textit{translation target} and the \textit{buggy code representation}. For the translation target, the model can predict the fixed code directly \citep{tufano2019empirical, phan2021cotext, yasunaga2021break} or generate some form of edit that can be applied to the buggy code to fix it \citep{yao2021learning, Chen2019, zhu2021syntax, yin2018learning}. 
For the buggy code representation, we can use a tokenized sequence of buggy code for sequence-to-sequence (Seq2Seq) prediction \citep{Chen2019, bhatia2018neuro} or program analysis representations such as abstract syntax tree (AST), data-flow graph (DFG), error messages  and build configurations \citep{yao2021learning, allamanis2021self, berabi2021tfix,tarlow2020learning}.




We propose NSEdit, a Transformer-based \citep{vaswani2017attention} model that predicts the editing sequence given only the source code. We use both encoder and decoder of the Transformer: the encoder processes the buggy code, and the decoder predicts the editing sequence given an edit grammar. We design the edit grammar as a regular language, and the Transformer uses it as a domain-specific language (DSL) to write scripts that can fix the bugs when executed. The grammar consists of two actions, \textit{delete} and \textit{insert}, which are added to the vocabulary of the language model as new tokens. The decoder has two modes: \textit{word/action mode} predicts the two action tokens along with word tokens, and \textit{location mode} selects the location of the edit.
A pointer network implements the location selection mode, and we slice the encoder memory as the embedding of the edit location to enable content-based retrieval, instead of representing a location as a static word embedding in the vocabulary.
We use beam search to generate predictions at inference time. Given diversity-versus-quality problem encountered in all beam-search-based sequence generation methods \citep{kool2019stochastic}, we train rerankers to improve the quality of sequences generated \citep{ng2019facebook, lee2021discriminative, ramesh2021zero}.
An overview of the architecture is provided in Figure \ref{fig:archi}. We will publish our source code.

We now introduce our decisions and hypotheses with respect to the two design factors of translation target and buggy code representations in the context of existing literature.

\paragraph{Translation target}
Using the fixed code as translation target is straight-forward to implement, with the added benefit that input and target are both code sequences, which can be easier to model and, compared to edits, more similar to the code corpus that large language models are pre-trained on.
However, as our results show, predicting the fixed code directly may encourage the model to learn the copying behavior which causes overfitting and does not reflect the goal of editing the code to make changes. 
Existing edit prediction approaches often rely on a graph representation of the buggy code \citep{yao2021learning}.
This implicitly assumes that there exists a graph representation for the buggy code in the first place, which may not be true if the bug causes syntactical errors.
Existing edit prediction approaches may also require significant architecture design and involve multiple stages of edit prediction  \citep{zhu2021syntax, hashimoto2018retrieve, yin2018learning}.
Instead, we design an edit grammar and rely on Transformer to do what it does best as a language model: learn the edit grammar and output editing instructions with this grammar.
As to the architectural changes, we add a pointer network to predict the location of an edit, which is an essential modification. Our use of rerankers is an orthogonal change that adds to the performance significantly but not the complexity of the main Transformer model.
The edit grammar can be seen as a DSL for edits, given which the Transformer generates a short program to edit the input sequence. This places our work alongside the burgeoning neural-symbolic literature to use neural networks to write executable scripts \citep{chen2021evaluating,mao2019neuro}, vastly expanding the algorithmic capacity of deep learning systems.


\paragraph{Buggy code representation}
We decide to train the model end-to-end given only the buggy code and fixed code without any auxiliary program analysis information. In light of recent finding that Transformers are universal sequence-to-sequence function approximators \citep{yun2019transformers}, we want to prove that Transformers are powerful enough to learn the syntax and data flow through pre-training on large code corpus and can do so robustly even when bugs are present in the input. When syntactical structures are not given, the Transformer model has to learn the syntax, which may provide additional signal that helps the model learn the semantics of the code corpus \citep{manning2020emergent}. When human programmers debug, they look at the code directly and only use AST/DFG as a mental model, and, as our results show, Transformer can learn to edit the code directly as well.
Representing buggy code with program analysis graphs can incorporate important static and dynamic analysis information. However, for code repair in particular, bugs may cause syntactical errors that prevent extraction of program analysis graphs. Program analysis tools may impose restrictions on the inputted programs (e.g. language-dependent tools), while our problem formulation is more general and portable. Lastly, it is a known challenge that neural program models encounter generalizability issues when semantic-preserving program transformations are encountered \citep{rabin2021generalizability}.

The main contributions of our paper are: 
(1) Our proposed method NSEdit achieves the state-of-the-art performance on the CodeXGLUE code repair benchmark \citep{lu2021codexglue}.  We show that pre-trained Transformer, given only the buggy code without program analysis representation or auxiliary information, can reach SOTA performance in code repair formulated as a sequence-to-sequence neural machine translation problem. 
(2) We formulate NSEdit grammar that is a regular language and one of the simplest edit representation in the literature. The two-mode decoder and finite state machine together ensure that the model follows the grammar. The Transformer uses the grammar as a neural-symbolic API to generate executable scripts that edit the code. We show that predicting editing sequences leads to superior performance, even when programs vary from packages to packages.
(3) We use a pointer network to achieve content-based edit location selection. We slice the encoder memory to obtain the latent representation of a potential edit location.
(4) We use an ensemble of rerankers to re-order the top-$K$ editing sequences produced by beam search and significantly improve all exact match accuracy. We apply a novel technique to fine-tune the rerankers on validation set which effectively reduces over-fitting of rerankers. 
(5) We show that the reranking score can be used to improve the precision of editing sequences with efficient trade-off of recall.




%


\section{Related work}

\paragraph {Code repair with deep learning} In addition to the code repair methods we discussed in the Introduction, we see a diverse array of methods proposed in recent years. 
Getafix \citep{Bader2019} presents a novel hierarchical clustering algorithm that summarizes fix patterns into a hierarchy and uses a simple ranking technique based on the context of a code change to select the most appropriate fix for a given bug. \citet{Vasic2019} presents multi-headed pointer networks to localize and fix the variable misuse bugs. Recently, \citet{Dinella2020} learns a sequence of graph transformations to detect and fix a broad range of bugs in Javascript: given a buggy program modeled by a graph structure, the model makes a sequence of predictions including the position of bug nodes and corresponding graph edits to produce a fix. DeepDebug \citep{drain2021deepdebug} trains a backtranslation Transformer model and uses various program analysis information obtained from test suites to fine-tune the model.



\paragraph {Neural machine translation with Transformer}  Transformer and its derivative models such as BERT \citep{devlin2018bert} and GPT \citep{radford2018improving} form a family of large language models that dominate neural machine translation and deep natural language processing. The models consists of many parameters, and the performance of the model scales with the size of the model \citep{brown2020language}. 
Recently, Transformer has been adapted to domains other than natural language processing, such as image classification \citep{dosovitskiy2020image} and protein structure modeling \citep{rao2021msa}, demonstrating Transformer as a general-purpose architecture. 




\section{Methods}


\paragraph{Problem setup}

Our code repair dataset contains pairs $(x, y)$ of strings, where $x$ is the source code that contains bugs (buggy code), and $y$ is the fixed code.  Given the buggy code as the input sequence $x$, the goal of NSEdit is to generate the correct sequence of edits $e$ that can transform $x$ into $y$.

For notations, we use variables such as $x$ to denote strings or constants. We use bold variables such as $\mathbf x$ to denote tokens and tensors. We use hatted variables such as $\hat{\mathbf x}$ or $\hat{x}$ to denote model predictions. We use uppercase bold variables such as $\mathbf X$ to denote a probability distribution.

\begin{figure}[t]
	\centering
	\includegraphics[width=9cm]{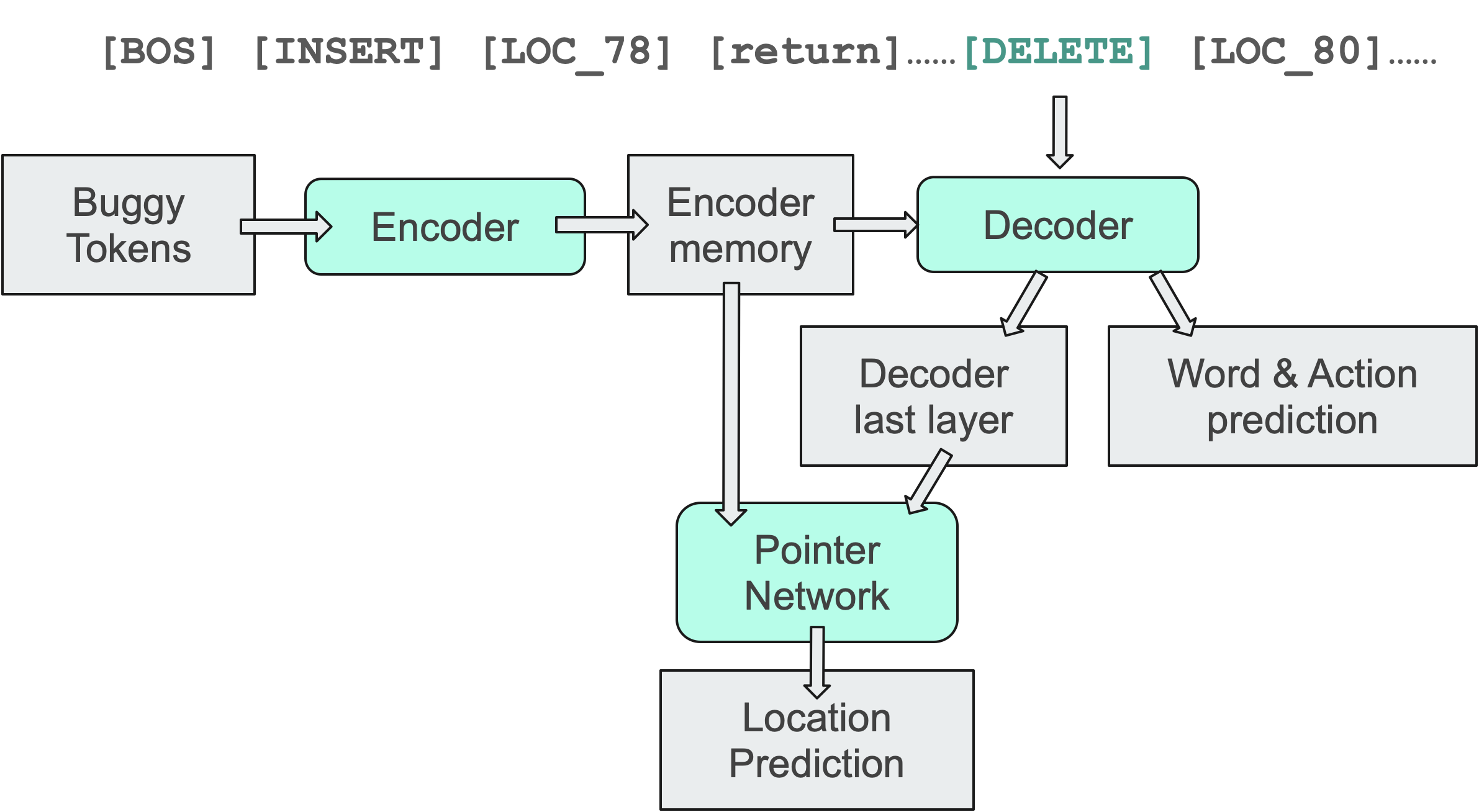}
	\caption{An illustration of the main NSEdit model architecture. There are two modes in the decoder. One mode predicts words and actions, and the other mode selects locations with a pointer network. The pointer network takes the penultimate layer output of the decoder and compares it with the encoder memory by dot product in order to select edit location.}
	\label{fig:archi}
\end{figure}



%
\paragraph{Overview of NSEdit Training}

Training NSEdit consists of a pipeline of  stages. We tokenize the buggy-fixed pair $(x, y)$ with pre-trained CodeBERT tokenizer before obtaining ground truth edits $e$. We load the pre-trained CodeBERT encoder \citep{feng2020codebert} and CodeGPT decoder \citep{lu2021codexglue}. We fine-tune the NSEdit model $f$ to predict editing sequences with teacher forcing. After training the model, we use beam search to generate the top-5 editing sequences (hypotheses). We train two rerankers with different architectures to classify which editing sequence is correct among the beam search hypotheses. The two rerankers and the original beam search score are combined with an ensemble model to produce the final reranking. Lastly, we fine-tune the rerankers on the beam search hypotheses on the validation set to reduce over-fitting. The beam search hypotheses reranked by the fine-tuned ensemble are the final predictions.


\subsection{Tokenization and editing sequence generation}

We tokenize the both buggy code and fixed code with a Byte Pair Encoding (BPE) \citep{sennrich2015neural} tokenizer that is pre-trained on CodeBERT code corpus. 
To compute the editing sequences, we use a variation of Ratcliff-Obershelp algorithm \citep{ratcliff1988pattern} implemented in Python's difflib library.
The editing sequences are computed on tokenized sequences instead of raw strings.

\begin{figure}[h]
	\centering
	\resizebox{9cm}{!}{%
		\begin{tikzpicture}[main/.style = {draw, circle}, node distance=3cm] 
			
			\node[main] (1) {ACTION}; 
			\node[main] (0) [left of=1] {BOS}; 
			\node[main] (3) [right of=1] {INS AT}; 
			\node[main, double] (2) [above of=3] {EOS}; 
			\node[main] (4) [below of=3] {DEL FROM}; 
			\node[main] (5) [right of=3] {WORD}; 
			\node[main] (6) [left of=4] {DEL TO}; 
			\node[ ] (invis) [left of = 0, xshift=1cm]{}; 
			
			\draw[-{Latex[scale=1.5]}] (0) -- node[midway, above] {[BOS]} (1);
			\draw[-{Latex[scale=1.5]}] (1) -- node[midway, above left, sloped, pos=0.65] {[EOS]} (2);
			\draw[-{Latex[scale=1.5]}] (1) -- node[midway, above] {[INSERT]} (3);
			\draw[-{Latex[scale=1.5]}] (1) -- node[midway, above right, sloped,  pos=0.1] {[DELETE]} (4);
			
			\draw[-{Latex[scale=1.5]}] (4) -- node[midway, above] {[LOC $l$]} (6);
			\draw[-{Latex[scale=1.5]}] (6) -- node[midway, left, pos=0.5] {[LOC $l$]} (1);
			
			\draw[-{Latex[scale=1.5]}] (3) to[bend left=10] node[midway, above] {[LOC $l$]} (5);
			
			\draw[-{Latex[scale=1.5]}] (5) --node[midway, below left, sloped, pos=0.1] {[DELETE]} (4);
			\draw[-{Latex[scale=1.5]}] ( 5) to[bend left=10]  node[midway, below] {[INSERT]} (3);
			\draw[-{Latex[scale=1.5]}] (5) -- node[midway, above right, sloped, pos=0.7] {[EOS]} (2);
			\draw[-{Latex[scale=1.5]}] (5) to[in=50,out=-50,loop] node[midway, right] {[\textit{w}]}(5);
			
			\draw[-{Latex[scale=1.5]}] (invis) -- (0);
			
		\end{tikzpicture} 
	}
	\caption{The transition diagram of the finite state machine for the NSEdit grammar used to generate editing sequences. The start state is the state BOS. The accept state is the state EOS.}
	\label{fig:grammar}
\end{figure}
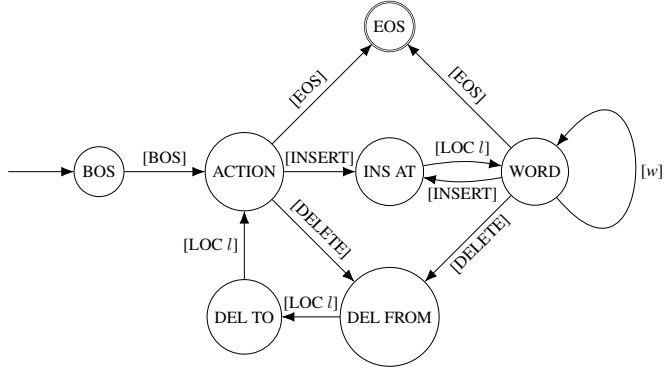

To formulate the NSEdit grammar formally, an editing sequence consists of two types of actions: $\textit{delete}(i,j)$ and $\textit{insert}(i, \mathbf s)$. The $\textit{delete}(i,j)$ action deletes the subsequence in $[i,j)$ from the buggy sequence $\mathbf x$. The $\textit{insert}(i, \mathbf s)$ action inserts a sequence of tokens $\mathbf s$ before location $i$ in the buggy sequence $\mathbf x$. 
As a result, NSEdit grammar contains three types of tokens in an editing sequence: action, word and location tokens.
The finite state machine that describes NSEdit grammar is provided in Figure \ref{fig:grammar}. An example editing sequence can be  \texttt{[DELETE] [LOC\_1] [LOC\_2] [INSERT] [LOC\_2] @Override public}.
More example editing sequences are provided in Appendix \ref{appendix:bugs}.

\subsection{Training Transformer to predict bug fix editing sequences with teacher forcing}
We use the Transformer model \citep{vaswani2017attention} to perform sequence-to-sequence prediction.
The NSEdit model computes $f(\mathbf x)=\hat{\mathbf e}$, where $\mathbf x$ is the buggy token sequence and $\hat{\mathbf e}$ is the predicted editing sequence. 
The encoder processes buggy code $\mathbf x$ and outputs the encoder memory $\mathbf m$, formally shown in Equation \ref{eq:encoder}. For input $\mathbf x$ with $L$ tokens and model with $h$ hidden units, the encoder memory has shape $(L, h)$, omitting the batch dimension. 
The decoder takes $\mathbf m$ and the current editing sequence token $\mathbf{e}_i$ as the input and autoregressively predicts the next token $\hat{\mathbf{e}}_{i+1}$ by maximum likelihood, as shown in Equation \ref{eq:autoregressive} and \ref{eq:argmax}, where $[\cdot]$ denotes the slicing operator in Python.
\begin{align}
	\mathbf{m} & = \textit{encoder}(\mathbf{x}) \label{eq:encoder}\\
	\hat {\mathbf{E}}_{i+1} & = \textit{decoder} (\mathbf{m}, {\mathbf{e}}_{i}) \label{eq:autoregressive}\\
	\hat {\mathbf e}_{i+1} & = \argmax_{w\in W}{\hat {\mathbf{E}}_{i+1}[w]} \label{eq:argmax}
\end{align}

We use teacher forcing as the training procedure \citep{williams1989learning,lamb2016professor}. This means that in Equation \ref{eq:autoregressive}, the ground truth edit token $\mathbf e_i$ is inputted into the decoder, but not the predicted token ${\hat {\mathbf e}}_i$.
For Seq2Seq models, teacher forcing decouples prediction $\hat {\mathbf{e}}_{i}$ from $\hat {\mathbf{e}}_{i+1}$ during back propagation, thus it is more robust against vanish/exploding gradient problems common in recurrent neural networks.

We fine-tune pre-trained CodeBERT \citep{feng2020codebert} and CodeGPT \citep{lu2021codexglue}. We modify the decoder to have two modes, a \textit{word/action mode} that predicts edit actions and inserted words, and a \textit{location mode} that predicts edit locations.

The original CodeBERT tokenizer has 50265 word tokens in the vocabulary, and we add \texttt{[DELETE]} and \texttt{[INSERT]} tokens to the vocabulary. When predicting words or actions, the decoder outputs a probability vector $\hat {\mathbf w}$ over a set $W$ of 50267 elements by passing the logits output $\mathbf c$ into the softmax function, shown in Equation \ref{eq:wordpred}.
\begin{align}
	\hat{\mathbf W}=\frac{\textit{exp}(\mathbf c)}{\sum_{j\in W}\textit{exp}(\mathbf c_i)} \label{eq:wordpred}
\end{align}

When predicting locations, instead of further expanding the vocabulary to add 513 location tokens and predict them along with words and actions, the decoder uses a pointer network in place of the last layer of the decoder (Figure \ref{fig:archi}).


The pointer network is a feed forward neural network. It transforms the output from the penultimate layer of the decoder into a latent representation $\mathbf v$ \citep{vinyals2015pointer,vasic2019neural}. In order to determine the location of the edit, we compute the dot product between $\mathbf v$ and $\mathbf m$ before a softmax function over all edit locations, as shown in equation \ref{eq:pointernetwork}. As the result, the pointer network outputs a probability vector $\hat {\mathbf L}$ over all edit locations at index $0,1,2...L$ for a buggy code with $L$ tokens.
\begin{align}
	\hat{\mathbf L}=\frac{\textit{exp}(\mathbf v^T \mathbf m)}{\sum_{j=0}^{L}\textit{exp}(\mathbf v^T \mathbf m_j)} \label{eq:pointernetwork}
\end{align}
Since ground truth is available with teacher forcing, we determine which decoder mode to use given the type of ground truth token $\mathbf e_{i+1}$.


We slice the encoder memory as the embedding $\mathbf m[l]$ to replace the embedding of a location token $\texttt{[LOC $l$]}$ as the input to the decoder in Equation \ref{eq:autoregressive}. 
As the result, the input $\mathbf m[l]$ and output $\mathbf v$ of the decoder for locations are both content-based representations, rather than a fixed location embedding that does not change when location context changes with the input program. 
We use cross entropy loss for both word/action prediction and location prediction and add them together with equal coefficients.


\subsection{Generating beam search hypotheses during inference}

\label{sec:beam search}

During inference, when ground truth is not available, we generate sequence predictions with beam search \citep{reddy1977speech,graves2012sequence,sutskever2014sequence}. Every partially generated sequence is assigned a probability $\Pi$ that is the product of every token probability, and the Top-$K$ most probable editing sequences (hypotheses) are outputted, where $K=5$. Formal definitions of our beam search procedure is provided in Appendix \ref{appendix:beam}.

The finite state machine can uniquely determine the next token type based on the previous token given the transition function in Figure \ref{fig:grammar}. Formally, we modify Equation \ref{eq:autoregressive} to use $\textit{fsm}$ to determine the token type as an input to the decoder
\begin{align}
	\hat {\mathbf{E}}_{i+1} = \textit{decoder}\big(\mathbf{m}, \hat{\mathbf e}_i, \textit{fsm}(\hat e_{i})\big)  \label{eq:inference}
\end{align}
We mask the probability of an invalid token to be zero, thereby ensuring valid NSEdit grammar syntax. We slice the encoder memory based on the predicted edit location $\hat {\mathbf e}_{i}$ during inference. Formally, when the current input is an edit location, $\hat{\mathbf e}_i = \argmax_{j=0,1,2..L}{\hat {\mathbf{E}}_{i}[j]} = \texttt{[LOC $l$]} = \mathbf m[l]$ in Equation \ref{eq:inference}.




\subsection{Reranking the beam search hypotheses}
\label{sec:reranker}
Our results show that top-5 accuracy is significantly higher than top-1 accuracy (Table \ref{tab:reranker ablations}), meaning that the correct edit can be produced among the 5 beam search hypotheses but not ranked the most probable by the original beam search probability $\Pi$ in Equation \ref{eq:bigpi}. Beam search with models trained with teacher-forcing can produce a diverse set of hypotheses, and the quality of the hypotheses may be improved \citep{kool2019stochastic}. To do so, we rerank the beam search hypotheses with rerankers \citep{ng2019facebook, lee2021discriminative}. We formulate this problem as a classification problem: given $K$ hypotheses produced by the beam search that have a correct prediction, the objective of the reranker is to classify which of the $K$ hypotheses is the correct one. The reranking score for a hypothesis $\hat {\mathbf e}_k$ is computed as
\begin{align}
	\textit{score}(\hat {\mathbf e}_k) =\log \frac{\exp(\textit{reranker}(\hat{\mathbf e}_k)/T)}{\sum_{i=1,2,..K}\exp(\textit{reranker}(\hat{\mathbf e}_i)/T)}
\end{align}
where $T=0.5$ is a temperature term that controls the smoothness of the probability distribution \citep{lee2021discriminative}. Note that each hypothesis $\hat{\mathbf e}_k$ goes through the same \textit{reranker} function where each  $\textit{reranker}(\hat{\mathbf e}_k)$ has a scalar output. This model can be seen as a special case of Siamese model \citep{koch2015siamese}.  We use cross entropy loss to train rerankers on the beam search hypotheses that are produced on the training set by the main NSEdit model.



We train two rerankers with different architectures: one with both Transformer encoder and decoder, the same architecture as the main NSEdit model, and the other with encoder only. The Transformer reranker outputs reranking score at the end of the sequence, and the encoder-only reranker outputs at the \texttt{[BOS]} token. Formally, for a beam hypotheses $\hat {\mathbf e}$ of length $L$, we compute the reranking scores 
\begin{align}
	\textit{reranker}_{\textit{Transformer}} (\hat {\mathbf e}) & = \textit{ff}_1(\hat{\mathbf E}_{L-1})\\
	\textit{reranker}_{\textit{Encoder}} (\hat {\mathbf e}) & = \textit{ff}_2(\mathbf m[\texttt{BOS}])
\end{align}
where $\hat {\mathbf E}_{L-1}$ is computed given by Equation \ref{eq:inference} and $\mathbf m$ by Equation \ref{eq:encoder}. The \textit{ff} function is a simple one-layer feed-forward neural network.

For every beam search hypotheses $\hat {\mathbf e}$, we have three ranking scores: the original beam search log probability score $\Pi$ (Equation $\ref{eq:bigpi}$) and two reranking scores. We use a linear ensemble model to blend the ranking scores: 
\begin{align}
	s= \log \Pi+ c_1 \textit{score}_{\textit{Transformer}}+ c_2 \textit{score}_{\textit{Encoder}} \label{eq:ensemble}
\end{align}
where $c_1, c_2$ are hyperparameters to be tuned \citep{jahrer2010combining,breiman1996bagging}. To search for the best hyperparameters, we train the rerankers on the training set, and we pick the configuration with the highest validation accuracy. 

\begin{table*}[ht]
		\caption{Top-1 exact match accuracy of NSEdit and other code repair models, evaluated on the \citet{tufano2019empirical} datasets. NSEdit achieved the state-of-the-art result on Tufano small abstract dataset, a part of CodeXGLUE benchmark. The results from CodeXGLUE benchmark or original papers are marked with $\dagger$. Tufano abstract dataset normalizes variable names, method names and type names. The best and second best results are bold and underlined.}
\label{tab:sota}
	\centering
	\resizebox{\linewidth}{!}{%
		\begin{tabular}{ c c rrrrrr }
			\toprule
			Length & Normalization& \textbf{NSEdit (ours)} & Baseline & CodeBERT$\dagger$ & GraphCodeBERT$\dagger$ & PLBART$\dagger$  & CoTexT$\dagger$\\ 
			\midrule
			Small  & Abstract & \textbf{24.04} & 16.30 & 16.40 & 17.3 & 19.21 & \underline{22.64}\\ 
			Small & Concrete & \textbf{23.86} &  \underline{17.75}  & - & - & - & - \\ 
			Medium & Abstract & \underline{13.87}           &   8.91   & 5.16  & 9.1 & 8.98 & \textbf{15.36}\\ 
			Medium & Concrete & \textbf{13.46} & \underline{9.59} & - &  - & - & -  \\
			\bottomrule
		\end{tabular}
	}
\end{table*}

Results show that rerankers tend to overfit on the training set. To mitigate reranker's overfitting issue, we fine-tune the rerankers on the validation set for $b$ epochs \citep{tennenholtz2018train}. 
To tune hyperparameter $b$, we re-split the validation set by 75:25, fine-tune the reranker on the $75\%$ split for $b$ epochs, and pick the $b$ with the highest accuracy on the $25\%$ split. We see that tuning for one epoch ($b=1$) performs the best. To produce the final fine-tuned ensemble reranker, we fine-tune both rerankers on 100\% of the validation set for one epoch, and combine them with the same ensemble hyperparameters $c_1, c_2$ found before fine-tuning.

\section{Experiments and Results}


\subsection{NSEdit achieves SOTA performance on CodeXGLUE code repair benchmark}

\label{sec:sota}

NSEdit achieved the state-of-the-art (SOTA) performance (24.04\%) on the \citet{tufano2019empirical} code repair dataset as a part of the CodeXGLUE benchmark \citep{lu2021codexglue}. We report our results on Tufano datasets in comparison with other code repair methods \citep{feng2020codebert, guo2020graphcodebert, ahmad2021unified, phan2021cotext} currently on the CodeXGLUE benchmark in Table \ref{tab:sota}. Compared to other methods, our method NSEdit is the only one that predicts any form of edits, while all others predict fixed programs directly. Some example bug fixes correctly produced by our NSEdit model are provided in Appendix \ref{appendix:bugs}. 

In Table \ref{tab:sota}, the \textit{Baseline} model is a Transformer with the CodeBERT encoder and a randomly initialized six-layer Transformer decoder. Compared to the complete NSEdit model, Baseline has four differences: (1) the prediction target is fixed code by default. (2) the decoder does not have a pointer network for location mode. (3) CodeGPT is not used to initialize the decoder. (4) rerankers are not used. We will reuse this Baseline model in the following experiments. 



\subsection{Predicting editing sequences performs better than predicting fixed code}

We formulate a novel NSEdit grammar to predict editing sequences as the target, rather than the fixed code. To confirm that predicting editing sequences yields better performance, we initialize two Baseline models without CodeGPT and rerankers, the same as in Section \ref{sec:sota}. One Baseline model predicts editing sequences and the other predicts fixed code. The only difference in the architecture is the pointer network needed to predict locations in editing sequences. We report the results on Tufano abstract datasets in Table \ref{tab:edit}. Predicting edits has better performance, possibly because editing sequence is shorter than fixed code, and it discourages copying behavior by focusing on the changes.

\begin{table}[ht]
	\begin{center}
			\caption{Exact match accuracy of two Baseline models when predicting editing sequences and fixed code on Tufano abstract dataset. The Baseline model is the main NSEdit model without CodeGPT and rerankers in order to isolate the effect of translation target.}
		\label{tab:edit}

			\begin{tabular}{ c c r r }
				\toprule
				Translation target & Length   & Top-1 & Top-5 \\ 
				\midrule
				\multirow{ 2}{*}{\textbf{Editing sequences}}     & Small	     &   \textbf{21.17} & \textbf{37.93} \\  
				& Medium    &  \textbf{13.20}  & \textbf{19.17}  \\ 
				
				\midrule
				\multirow{ 2}{*}{Fixed code}	 & Small	     & 16.30 & 30.42 \\ 
				& Medium    & 8.91   & 17.14  \\ 
				
				\bottomrule
			\end{tabular}

	\end{center}
\end{table}

\subsection{NSEdit performs robustly against package-to-package variations}

We note that the \citet{tufano2019empirical} training, validation and test sets have overlapping Java packages. 
To investigate the effect of package-to-package variations, we curate an in-house dataset from the publicly available 10K Github Java packages \citep{githubCorpus2013}, which will be public. Our dataset generation process resembles \citet{tufano2019empirical}: we partition the dataset given the buggy program length and normalize variable names, method names and type names in abstract code, while retaining the original names in concrete code. We implement a strict policy to separate training, validation and test set packages. Other than the same packages, closely related packages that share same naming prefix, e.g. ``spring-cloud-stream-samples'' and ``spring-cloud-stream'', are considered related and also separated in either training, validation or test set. The dataset consists of 138575/12983/9282 train/valid/test samples. We report the accuracy of Baseline models in Table \ref{tab:house}. 

\begin{table}[ht]
	\begin{center}
			\caption{Exact match accuracy of Baseline models that predict on editing sequences and fixed code on our in-house dataset (small). Mixed dataset mixes the training and validation set packages and leave the test set separate, and otherwise all three sets have separate packages. We see that when packages are separate or when input programs have original variable names, it overfits less to predict editing sequences than fixed code directly. CodeGPT and rerankers are not used.}
		\label{tab:house}

			\begin{tabular}{ c c c r r r r }
				\toprule
				Packages  & Norm.  &  Target & \multicolumn{2}{c}{Top-1} & \multicolumn{2}{c}{Top-5} \\ 
				&    &  & Val. & Test & Val. & Test \\ 
				\midrule

				\multirow{ 4}{*}{Mixed}        & \multirow{ 2}{*}{Abs.}     &  Code   & \textbf{26.38} & \textbf{9.78}  & \textbf{41.67}  & \textbf{21.99} \\ 
				&    & Edit   &  26.14  & 9.07 & 38.63  &  18.51 \\  
				\cmidrule{2-7}
				& \multirow{ 2}{*}{Conc.}    & Code&  27.61 & 3.45 & 37.45 & 8.77 \\ 
				&    & Edit & \textbf{30.91} & \textbf{7.11}  & \textbf{38.52} &   \textbf{12.11} \\ 
				\midrule
				\multirow{ 4}{*}{Separate}    & \multirow{ 2}{*}{Abs.}   & Code & \textbf{12.72} & 10.82 & \textbf{27.50} & 26.45 \\  
				&      &  Edit  & 12.51 & \textbf{11.46}  & 27.03 & \textbf{27.05}  \\ 
				\cmidrule{2-7}
				& \multirow{ 2}{*}{Conc.} & Code&  4.94 & 3.98 & 11.93 & 11.15 \\ 
				& &  Edit  &\textbf{9.10} &\textbf{9.04} &\textbf{16.30} & \textbf{18.42} \\ 
				\bottomrule
			\end{tabular}

	\end{center}
\end{table}

Recall that we hypothesize previously that predicting fixed code directly encourages the model to learn the copying behavior.
We see that when packages are mixed, all models overfit with large gap between validation and test accuracy.
Furthermore, mixing packages causes all models to have reduced accuracy, likely because overfitting is a significant performance bottleneck.
When predicting code, accuracy on concrete code (27.61\%, row 3) is slightly better than accuracy on abstract code (26.38\%, row 1)  on validation set with mixed training/validation packages, but only half at test time with unseen packages (3.45\% v.s. 9.78\%), possibly because for concrete code, the model is given more context and it is easier to copy from similar programs, which makes overfitting more severe. 
Predicting edits does not suffer the same performance drop at test time for concrete code (7.11\%, row 4) compared to abstract code (9.07\%, row 2), which supports our hypothesis that predicting edits discourages copying behavior.
When packages are separate and code is concrete, predicting editing sequences (9.04\%, row 8) doubles the test time top-1 accuracy of predicting fixed code directly (3.98\%, row 7).
This is the most valuable use case in applications, and predicting edits has even greater advantage than predicting fixed code.


\subsection{NSEdit  is a general bug fix method in different languages and settings}

The NSEdit grammar is language agnostic. To confirm that NSEdit works on languages other than Java, we evaluate our NSEdit model on ETH Py150 dataset \citep{kanade2020learning, raychev2016probabilistic}. The Python dataset contains five classification tasks, and we experiment on three of them that are related to code repair: variable-misuse classification,  wrong binary operator, and swapped operand. We process ETH Py150 according to our problem setup and report the performance of NSEdit in Table \ref{tab:py}. We confirm that NSEdit is a general method can edit buggy programs in different languages and settings.

\begin{table}[t]
	\centering
		\caption{Exact match accuracy of NSEdit on three code repair tasks of the ETH Py150 dataset. NSEdit can repair programs in different languages and settings.} 
	\label{tab:py}

		\begin{tabular}{ c rr }
			\toprule
			Task & Top-1 & Top-5 \\ 
			\midrule
			\multirow{ 1}{*}{Variable-misuse classification}     &  83.08 & 87.14 \\
			\multirow{ 1}{*}{Wrong binary operator}                  &  58.44 &  75.32 \\
			\multirow{ 1}{*}{Swapped operand}                          & 67.38   &  69.23  \\
			\bottomrule
		\end{tabular}

\end{table}


\subsection{Pre-trained encoder and decoder can be fine-tuned together on editing sequences }

Fine-tuning pre-trained models on code repair contributes to the performance of our model. To investigate the effect of different backbone models, we change the pre-trained backbones used to initialize the weights of the Transformer before fine-tuning. We report accuracy on Tufano datasets in Table \ref{tab:bone}.

\begin{table}[h]
	\centering
		\caption{Exact match accuracy of NSEdit when different pre-trained backbone models are used to initialize the weights before fine-tuning on Tufano abstract dataset. The pre-trained encoder and decoder can be fine-tuned together in the same model. The decoder, despite not pre-trained on editing sequences, can be fine-tuned to predict edits. Reranking is not performed to isolate the effect. All models predict editing sequences.}
	\label{tab:bone}

		\begin{tabular}{ c c r r r }
			\toprule
			Length & Pre-trained backbone & Top-1 & Top-3 & Top-5 \\ 
			\midrule			
			\multirow{ 3}{*}{Small} & No backbone   & 15.89 & 25.00 & 27.66 \\ 
			& CodeBERT &  21.17 & 33.40 & 37.93  \\ 
			&  \textbf{CodeBERT+CodeGPT}  & \textbf{22.35} & \textbf{33.20} & \textbf{36.35}  \\ 
			\midrule
			\multirow{ 3}{*}{Medium}  & No backbone &  7.32  & 10.89 & 11.81 \\ 
			&	CodeBERT  & 13.20 & 17.68 & 19.17 \\ 
			&	\textbf{CodeBERT+CodeGPT}   &  \textbf{13.72} & \textbf{18.87} & \textbf{20.17}\\ 
			\bottomrule
		\end{tabular}

\end{table}
%
%

The CodeBERT encoder backbone is trained on multiple programming languages, including Java. Even when the input sequences contain bugs and program analysis auxiliary information is not provided, CodeBERT can robustly extract program information and improve the performance of NSEdit after fine-tuning. The CodeGPT decoder backbone also improves the performance of NSEdit. CodeBERT and CodeGPT are two independently published models pre-trained on different datasets, and our results confirm that even when large language models are pre-trained in different settings, they can be integrated into the same Transformer model. Notably, CodeGPT is pre-trained on code, and the model can effectively transfer knowledge to predict editing sequences. 

\subsection{Rerankers bring significant performance improvements}
The use of rerankers significantly improves the performance of NSEdit to the state-of-the-art level. We investigate the effect of different settings of rerankers in Table \ref{tab:reranker ablations}. 

\begin{table}[ht]
		\caption{Top-5 exact match accuracy of incremental ablations in reranker settings on Tufano small abstract dataset. Ensemble improves all accuracies compared to a single Transformer reranker. Fine-tuning on validation set improves all accuracies as well. The combination of both achieves the highest accuracy.}
	\label{tab:reranker ablations}
	\begin{center}

			\begin{tabular}{ c rrrrr}
				\toprule
				Reranker settings & Top-1 & Top-2 & Top-3 & Top-4 & Top-5 \\ 
				\midrule
				NSEdit without rerankers & 22.35 & 29.19 & 33.20 & 35.22  & 36.35 \\ 
				Transformer & 18.47 &  26.80 &  31.55 & 34.62 & 36.35 \\
				Ensemble & 22.76 & 29.07 & 32.63 & 35.08 & 36.35 \\
				Fine-tuned Transformer  & 21.17  & 28.67 & 32.84 & 35.34 & 36.35 \\
				\textbf{Fine-tuned ensemble}  & \textbf{24.04} & \textbf{30.54} & \textbf{34.10} & \textbf{35.70} & \textbf{36.35} \\
				\bottomrule
			\end{tabular}

	\end{center}
\end{table}

The final reranker ablation \textit{Fine-tuned ensemble} achieves the best accuracy among all settings and is presented in the Methods section. Notably, we see that the use of Transformer reranker has lower accuracy than without reranker. Ensemble rerankers and validation-set fine-tuning both improved performance separately and in combination. Further details about reranker ablation experiment settings are provided in Appendix \ref{appendix:reranker ablation details}.  

\subsection{Ensemble reranker efficiently trades off precision and recall}
\begin{figure}[t]
	\centering
	\includegraphics[width=10cm]{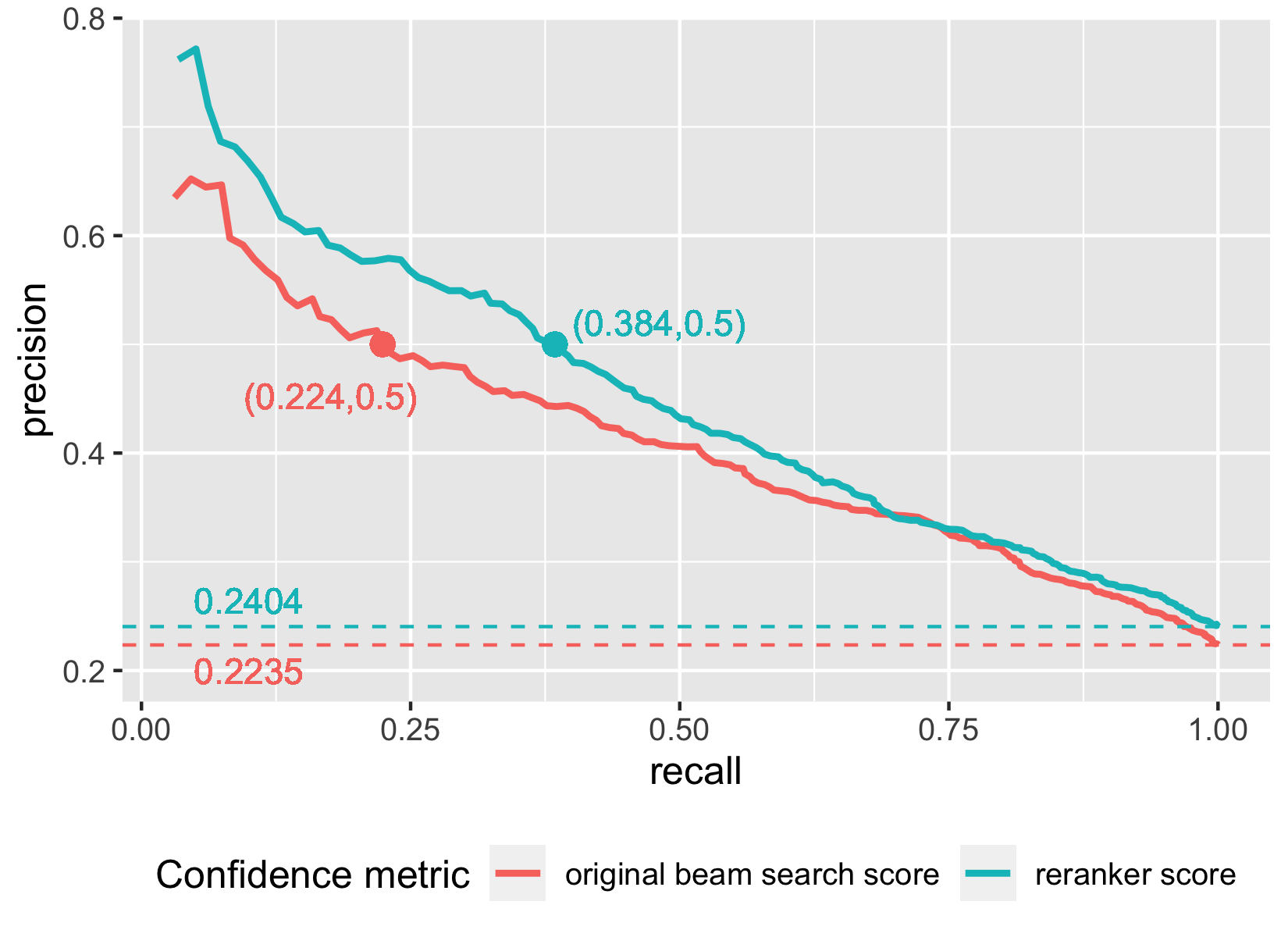}
	\caption{Precision versus recall plot on Tufano small abstract dataset using reranking score and original beam search score as the confidence metric. Among the two, reranking score trades off precision and recall more efficiently, reaching the same precision with higher recall.}
	\label{fig:pr}
\end{figure}

Rerankers rerank the beam search hypotheses as discussed in Section \ref{sec:reranker}. A beam search hypothesis has higher reranking score if it is predicted to be more likely. Therefore, we can use the reranking score computed in Equation \ref{eq:ensemble} as a confidence metric to trade off precision and recall of the editing sequence generated by beam search. If the reranking score is lower than a threshold, we discard the editing sequence predicted (negative), and otherwise we use it (positive). We sweep the threshold parameter and generate the precision-recall trade-off plot for Tufano small abstract dataset in Figure \ref{fig:pr}. Note that for accuracy reported previously, the model makes a prediction every time (always positive) and recall is 100\%.  

When the model is not required to make a prediction for every buggy sequence, precision can be improved with trade-off of recall by using the reranking score as a confidence metric. This trade-off is usually desirable in bug fix applications.
Specifically, when recall is lowered from 100\% to 38.4\%, precision can be increased from 24.04\% to 50\%. 
In comparison, we also use beam search's original sum of log probability score computed in Appendix Equation \ref{eq:bigpi} as the confidence metric, and we see that reranking score is more efficient because its recall is higher at every precision level. Specifically, the original score's recall needs to be lowered to 22.4\% in order to reach the same precision level of 50\%, which is almost half of the reranking score's recall. 
This result additionally confirms that our rerankers learn an extrinsic signal that differs from the main model's intrinsic confidence.

%

\section{Discussions}
NSEdit has achieved a new state-of-the-art performance on the code repair task of the CodeXGLUE benchmark \citep{lu2021codexglue, tufano2019empirical}. For code repair, it is more effective to predict editing sequences than fixed code.

As closing thoughts, we want to draw attention that our edit grammar generates a neural-symbolic interface that is a special case of a general paradigm. Integration of deep learning systems with traditional symbolic systems can be achieved by formulating a domain-specific language (DSL) and training a Transformer to use the DSL as a programmable interface.



%

\bibliography{my}
\bibliographystyle{iclr2022_conference}

\appendix

\section{Beam search formal definition}
\label{appendix:beam}

During beam search, we maintain $K$ incomplete subsequences (i.e. beams) $B_{i,k}$, where  $B_{0,k}=[\texttt{[BOS]}]$ for all $k=1,2...K$. For any $B_{i,k}$, we compute the probability $\Pr (\hat e_{i+1,k} | \hat e_{1,k}, \hat e_{2,k}, ... \hat e_{i,k}) = \hat {\mathbf E}_{i+1,k}[\hat e_{i+1,k}]$ given Equation \ref{eq:inference}. 
The probability of the concatenated sequence $B_{i+1,k}=B_{i,k}+[\hat e_{i+1,k}]=[\hat e_{1,k}, \hat e_{2,k}, ... \hat e_{i,k}, \hat e_{i+1,k}]$ is the product of the probabilities of all tokens as they are generated. Formally,
\begin{align}
	\Pr(\hat e_1, \hat e_2... \hat e_{i+1}) =  \prod_{j=1}^{i+1} & \Pr (\hat e_j | \hat e_1, ... \hat e_{j-1}) \label{eq:bigpi}
\end{align}
We denote this product as $\Pi$ in Equation \ref{eq:ensemble}.
Taking logarithm on both sides, we have
\begin{align}
	\log \Pr(B_{i,k}+[\hat e_{i+1,k}]) =  & \!\begin{aligned}[t] & \log \Pr(B_{i,k}) +\\ 
		&  \log \hat {\mathbf E}_{i+1,k}[\hat e_{i+1,k}]   \end{aligned}  
\end{align}

We take the top-$K$ most probable tokens $\hat e_{i+1,k}\in W$ for all beam $k=1,2,...K$, given 
\begin{align}
	\argtopk_{(k,\hat e_{i+1,k}) \in \{1,2,...K\} \times W} \log \Pr(B_{i,k}+[\hat e_{i+1,k}]) 
\end{align}
where any previous step beam may have multiple concatenated beams in top-$K$ with different tokens. The selected top-$K$ tokens $\hat e_{i+1,k}$ form a new set of $K$ beams for the next iteration, until \texttt{[EOS]} is predicted or maximum length is reached. Note that since $\argtopk$ is applied to $\{1,2,...K\} \times W$, it is possible for different tokens to append to the same previous beam and are all included in the top-$k$ beams for the next step. Also note that beam $B_{i+1,k}$ may not contain beam $B_{i,k}$ as a subsequence.

\section{Reranker ablation  experiments details}
\label{appendix:reranker ablation details}

In this section, we provide more details on the settings of the reranker ablation experiments presented in Table \ref{tab:reranker ablations}.

The \textit{NSEdit without rerankers} version does not use rerankers and directly reports beam search accuracy based on the original beam search score $\Pi$ in Equation \ref{eq:bigpi}. We compare the reranked accuracy with this baseline.

The \textit{Transformer} version trains a single Transformer reranker to rerank the beam search hypotheses, ignoring the original beam search score. We see that the accuracy is lower than the accuracy of \textit{NSEdit without rerankers}.

The \textit{Ensemble} version trains both rerankers and blends the three ranking scores by optimizing validation accuracy. We see that the ensemble reranker improves over \textit{Transformer} reranker but not better than \textit{NSEdit without rerankers}. 

The \textit{Fine-tuned Transformer} version trains a single Transformer reranker and tunes it on validation set for one epoch. We see that it significantly outperforms the \textit{Transformer} version without fine-tuning, which suggests that fine-tuning on validation prevents the reranker from overfitting. The accuracy of this ablation outperforms \textit{NSEdit without rerankers}.

The final version \textit{Fine-tuned ensemble} uses a blended ensemble of two rerankers fine-tuned on the entire validation set, as described in the Methods section.

\section{Training and model hyperparameters}

We implement our Transformer architecture in PyTorch, except the pre-trained CodeBERT and CodeGPT models, which we load from HuggingFace \citep{wolf2019huggingface}
. The learning rate is set to be 1e-4 multiplied by the number of GPUs. When CodeGPT weights are loaded, we halve the learning rate of pre-trained parameters and quadruple the learning rate of randomly initialized parameters. AdamW optimizer \citep{loshchilov2017decoupled} with triangular learning rate scheduler is used in all experiments. The NSEdit main model is trained for at most 60 epochs and early stopping is applied if the accuracy does not improve. Automatic mixed precision (AMP) is enabled. Training of the model together with rerankers on Tufano datasets takes around a day on a machine with 4 V100 Nvidia GPUs.

In beam search, a length penalized score is computed for the partially generated sequences at every step according to \citet{wu2016google}. After the scores are computed, the finite state machine set the invalid tokens to have zero probability according to the edit grammar as described in Section \ref{sec:beam search}.

Rerankers are trained on buggy programs for which the beam search produces at least one correct editing sequence. We train rerankers for 12 epochs, with the same learning rate setup as the main NSEdit model. We fine-tune the reranker on validation set for 1 epoch with 1 GPU.

We find the best hyperparameters with Ray tune \citep{liaw2018tune} or grid search. The ensemble reranker on Tufano datasets have coefficients reported in Table \ref{tab:ensemble}. Each coefficient is searched 
among 10 candidates in logarithmic interval $[0.01, 100]$, then another 20 candidates in a narrower linear interval $[0.1, 2]$.


\begin{table}[ht]
	\begin{center}
			\caption{The ensemble reranker hyper-parameters found through grid search for Tufano datasets. The ensemble reranker is a linear model with Equation \ref{eq:ensemble}. }
		\label{tab:ensemble}
			\begin{tabular}{ c c rr }
				\toprule
				Length  & Normalization& Transformer $c_1$ & Encoder $c_2$ \\ 
				\midrule

				\multirow{2}{*}{Small}      & Abstract   &   0.4  &                0.4 \\
				& Concrete   & 1.0 &                0.7 \\
				\midrule
				\multirow{2}{*}{Medium}  & Abstract &  0.2&                0.3 \\
				& Concrete &  0.3 &                0.1 \\
				
				\bottomrule
			\end{tabular}
	\end{center}

\end{table}


\section{Tufano dataset editing sequence statistics}
In Table \ref{tab:dataset}, we summarize statistics about the ground truth editing sequences in \citet{tufano2019empirical} datasets. We see that the editing sequences in Tufano medium, compared to Tufano small dataset, have more number of edits, longer insertion length and longer overall editing sequence length by mean and median. This suggests that the bug fixes in Tufano medium dataset are overall more difficult to predict correctly.

\begin{table}[ht]
	\begin{center}
			\caption{The statistics of the ground truth editing sequences on Tufano datasets. We see that the editing sequences from Tufano medium, compared to Tufano small dataset, have more number of edits, longer insertion length and longer overall editing sequence length by mean and median.}
		\label{tab:dataset}
		\begin{tabular}{cc rr}
			\toprule
			& Tufano   & Mean & Median \\
			\midrule
			\multirow{3}{2cm}{Number of edits}      & small    & 2.02 & 2      \\
			& medium   & 2.45 & 2      \\
			& combined & 2.24 & 2      \\
			\midrule
			\multirow{3}{2cm}{Insertion length}    & small    & 3.60 & 1      \\
			& medium   & 6.22 & 2      \\
			& combined & 4.99 & 1      \\
			\midrule
			\multirow{3}{2cm}{Editing sequence length}  & small    & 10.8 & 8      \\
			& medium   & 14.4 & 10     \\
			& combined & 12.7 & 8     \\
			\bottomrule
		\end{tabular}
	\end{center}
\end{table}

\section{Example bug fixes}
\label{appendix:bugs}

We provide some examples of bug fix that are correctly produced by NSEdit in Figures \ref{fig:firstbug} to \ref{fig:lastbug}.

\begin{figure*}[h]
\begin{lstlisting}[style=diffjava]
- public void write(byte b[]) throws IOException {
?                         --
+ public void write(byte[] b) throws IOException {
?                       ++
    assertOpen();
    super.write(b);
  }
\end{lstlisting}
\cprotect\caption{Example bug fix. Coding style improvement. The predicted editing sequence is \Verb|[INSERT][LOC_6][][DELETE][LOC_7][LOC_9][INSERT][LOC_9])| }
\label{fig:firstbug}
\end{figure*}

\begin{figure*}[h]
\begin{lstlisting}[style=diffjava]
  @Override public Iterator<?> downstreams(){
    WindowGroupedFlux<T> g=window;
+   if (g == null) {
-   if (g == null)   return Collections.emptyList().iterator();
?  ---------------
+     return Collections.emptyList().iterator();
+   }
    return Collections.singletonList(g).iterator();
  }
\end{lstlisting}
\cprotect\caption{Example bug fix. Coding style improvement. The predicted editing sequence is \Verb|[DELETE][LOC_31][LOC_34][INSERT][LOC_34] {return[INSERT][LOC_41]}|}
\end{figure*}

\begin{figure*}[h]
\begin{lstlisting}[style=diffjava]
  public void addPoint(Point2D point){
    ArcPoint newPoint=new ArcPoint(point,false);
-   HistoryItem historyItem=new AddArcPathPoint<S,T>(arc,newPoint);
?                                               ---
+   HistoryItem historyItem=new AddArcPathPoint<>(arc,newPoint);
    historyItem.redo();
    historyManager.addNewEdit(historyItem);
  }
\end{lstlisting}
\cprotect\caption{Example bug fix. Coding style improvement. The predicted editing sequence is \Verb|[DELETE][LOC_37][LOC_40][DELETE][LOC_64][LOC_66]|}
\end{figure*}

\begin{figure*}[h]
\begin{lstlisting}[style=diffjava]
  public long getConsoleReportingInterval(){
-   System.out.println(reportingIntervalConsole.getValue());
    return reportingIntervalConsole.getValue();
  }
\end{lstlisting}
\cprotect\caption{Example bug fix. Remove unnecessary logging statement. The predicted editing sequence is \Verb|[DELETE][LOC_9][LOC_24]|}
\end{figure*}

\begin{figure*}[h]
\begin{lstlisting}[style=diffjava]
- public String getName(){
+ @Override public String getName(){
? ++++++++++
    return CypherPsiImplUtil.getName(this);
  }
  
\end{lstlisting} %
\cprotect\caption{Example bug fix. Missing annotation. The predicted editing sequence is \Verb|[DELETE][LOC_1][LOC_2][INSERT][LOC_2]@Override public|}
\end{figure*}

\begin{figure*}[h]
\begin{lstlisting}[style=diffjava]
  @Override public boolean apply(PickleEvent pickleEvent){
    String picklePath=pickleEvent.uri;
    if (!lineFilters.containsKey(picklePath)) {
-     return true;
?            ^^^
+     return false;
?            ^^^^
    }
    for (  Long line : lineFilters.get(picklePath)) {
      for (    PickleLocation location : pickleEvent.pickle.getLocations()) {
        if (line == location.getLine()) {
          return true;
        }
      }
    }
    return false;
  }
\end{lstlisting}
\cprotect\caption{Example bug fix. Logical error. The predicted editing sequence is \Verb|[DELETE][LOC_44][LOC_45][INSERT][LOC_45] false|}
\end{figure*}

\begin{figure*}[h]
\begin{lstlisting}[style=diffjava]
  /** 
   * Returns the preferred fragment size.
   * @param format target format
   * @return the preferred fragment size
   * @throws IOException if failed to compute size by I/O error
   * @throws InterruptedException if interrupted
   * @throws IllegalArgumentException if some parameters were {@code null}
   */
  public long getPreferredFragmentSize(FragmentableDataFormat<?> format) throws IOException, InterruptedException {
    if (format == null) {
      throw new IllegalArgumentException("format must not be null");
    }
    long min=getMinimumFragmentSize(format);
-   if (min <= 0) {
?            -
+   if (min < 0) {
      return -1;
    }
    long formatPref=format.getPreferredFragmentSize();
    if (formatPref > 0) {
      return Math.max(formatPref,min);
    }
    return Math.max(preferredFragmentSize,min);
  }
\end{lstlisting}
\cprotect\caption{Example bug fix. Logical error. The predicted editing sequence is \Verb|[DELETE][LOC_140][LOC_141][INSERT][LOC_141] <|}
\end{figure*}

\begin{figure*}[h]
\begin{lstlisting}[style=diffjava]
  public boolean equals(AudioQuality quality){
    if (quality == null)   return false;
-   return (quality.samplingRate == this.samplingRate & quality.bitRate == this.bitRate);
+   return (quality.samplingRate == this.samplingRate && quality.bitRate == this.bitRate);
?                                                      +
  }
\end{lstlisting}
\cprotect\caption{Example bug fix. Wrong operator. The predicted editing sequence is \Verb|[DELETE][LOC_35][LOC_36][INSERT][LOC_36] &&|}
\end{figure*}

\hspace*{-30cm}
\begin{figure*}[h]
\noindent
\begin{lstlisting}[style=diffjava]
  @Override public void run(){
    this.ownerThread=Thread.currentThread();
    Log.debug("Starting event loop","name",name);
    setStatus(LoopStatus.BEFORE_LOOP);
    try {
      beforeLoop();
    }
   catch (  Throwable e) {
-     Log.error("Error occured before loop is started","name",name,"error",e);
+     Log.error("Error occurred before loop is started","name",name,"error",e);
?                          +
      setStatus(LoopStatus.FAILED);
      return;
    }
    setStatus(LoopStatus.LOOP);
    while (status == LoopStatus.LOOP) {
      if (Thread.currentThread().isInterrupted()) {
        break;
      }
      try {
        insideLoop();
      }
   catch (    Throwable e) {
        Log.error("Event loop exception in " + name,e);
      }
    }
    setStatus(LoopStatus.AFTER_LOOP);
    afterLoop();
    setStatus(LoopStatus.STOPPED);
    Log.debug("Stopped event loop","name",name);
  }
\end{lstlisting}
\cprotect\caption{Example bug fix. Spelling error. The predicted editing sequence is \Verb|[DELETE][LOC_68][LOC_70][INSERT][LOC_70] occurred|}
\label{fig:lastbug}
\end{figure*}

\end{document}